\documentclass{article}

\usepackage{PRIMEarxiv}

\usepackage[utf8]{inputenc} 
\usepackage[T1]{fontenc}    
\usepackage{hyperref}       
\usepackage{url}            
\usepackage{booktabs}       
\usepackage{amsfonts}       
\usepackage{nicefrac}       
\usepackage{microtype}      
\usepackage{lipsum}
\usepackage{fancyhdr}       
\usepackage{graphicx}       
\graphicspath{{media/}}     

\usepackage{caption}
\usepackage{subcaption}
\usepackage{algorithm}
\usepackage{algpseudocode}
\usepackage{amsmath}
\usepackage{multirow}
\title{Attention and Autoencoder Hybrid Model for 
Unsupervised Online Anomaly Detection
}

\author{
  Seyed Amirhossein Najafi \\
  Department of Electrical and Computer Engineering \\
  Shiraz University \\
  Shiraz, Iran\\
  \texttt{amirhoseinnajafi46@gmail.com} \\
   \And
  Mohammad Hassan Asemani \\
  Department of Electrical and Computer Engineering \\
  Shiraz University \\
  Shiraz, Iran\\
  \texttt{asemani@shirazu.ac.ir} \\
\And
  Peyman Setoodeh\\
  Department of Electrical and Computer Engineering \\
  Shiraz University \\
  Shiraz, Iran\\
  \texttt{psetoodeh@shirazu.ac.ir} \\
}

\begin{document}
\maketitle

\begin{abstract}
This paper introduces a hybrid attention and 
autoencoder (AE) model for unsupervised online anomaly 
detection in time series. The autoencoder captures local 
structural patterns in short embeddings, while the attention 
model learns long-term features, facilitating parallel computing 
with positional encoding. Unique in its approach, our proposed 
hybrid model combines attention and autoencoder for the first 
time in time series anomaly detection. It employs an attention-based mechanism, akin to the deep transformer model, with key 
architectural modifications for predicting the next time step 
window in the autoencoder's latent space. The model utilizes a 
threshold from the validation dataset for anomaly detection and 
introduces an alternative method based on analyzing the first 
statistical moment of error, improving accuracy without 
dependence on a validation dataset. Evaluation on diverse real-world benchmark datasets and comparing with other well-established models, confirms the effectiveness of our proposed 
model in anomaly detection.
\end{abstract}


\section{Introduction}\label{sec1}
Anomaly detection for time series involves identifying 
unexpected and unfitting observations throughout time to 
enlighten us about abrupt changes in the data. There are many 
applications for anomaly detection in numerous fields, such as 
attack detection in cyber-physical systems [1], [2], sensor 
failure detection [3], Internet of Things (IoT) [4], energy 
engineering [5], and control systems [6]. It is critical to deploy 
unsupervised and online models, which are fast enough to 
perform anomaly detection at a pace corresponding to the rate 
of data generation.
The proposed model in this paper addresses this issue.
Therefore, it does not need labeled data to learn anomaly 
detection, as acquiring accurate enough labeled data is 
laborious and time-consuming. It is online in the sense that it 
can predict the existence of anomalies at each step without 
needing all the data to be available. Due to the optimized 
model and the attention-based mechanism for forecasting, the 
proposed method can work in a real-time manner using 
appropriate hardware depending on the data generation 
frequency. A number of accurate online anomaly detection 
algorithms have been proposed in the literature, such as the 
VAE-LSTM hybrid model [7], which uses recurrent neural 
networks (RNNs). Such algorithms receive time series with 
temporal order, and learn the temporal relations in the time 
series. However, they may not render themselves easily to 
parallel computation, and in effect therefore, they may be 
slow. The birth of transformers in 2017 was the start of a 
revolution that turned the odds against the RNNs [8]. Based 
on an attention mechanism initially used for Natural Language 
Processing (NLP), the transformer model allows for learning 
the temporal relation using positional embedding to speed up 
the process. Since its release, transformer has been used in 
time series forecasting applications and has shown outstanding 
performance [9].
There are three anomaly types [10]: point, collective, and 
contextual anomalies. If a point significantly deviates from the 
rest of the data, it is considered as a point anomaly, which is 
the most accessible and straightforward type to detect. In some 
cases, individual points are not anomalous, but a sequence of 
points is labeled as an anomaly; they are known as collective 
anomalies. Long-term information is needed to detect this 
anomaly type. Some events can be expected in a particular 
context while detected as anomalies in another context; these 
are categorized as contextual anomalies, and local information 
is needed for their detection. As it is clear, there is no 
inconvenience in the detection of point anomalies. However, 
on the contrary, in order to detect collective and contextual 
anomalies, there is a requirement to consider the local and 
long-term temporal relationships. This paper proposes an attention and autoencoder joint model as a reliable and fast 
anomaly detection model. It benefits from the autoencoder’s 
representation learning power as a deep generative model and 
the temporal modeling ability of the deep attention model.
Using the AE network, our model captures the structural 
regularities of the time series over local windows, and the 
attention model attempts to model longer-term trends. In 
summary:
\begin{itemize}
	\item	The AE network is used to capture the structural regularities of the time series over local windows and summarize the windows into short embeddings.
	\item	The attention-based network is trained to learn long-term temporal relations in the latent space of the AE.
	\item	The proposed merged structure allows for detecting all types of anomalies by taking account of both long- and short-term characteristics.
\end{itemize}
The rest of the paper is organized as follows. Section~2 discusses autoencoders and the transformer model, while Section~3 introduces the proposed model model. Section~4 demonstrates the experimental results and compares the proposed method with other well-established and commonly used models in the literature. Finally, Section~5 expresses the concluding remarks.

\section{Background \& Related Work}\label{sec2}
This section first reviews two neural network models; 
autoencoder and transformer, both of which are either used in 
our proposed model or have inspired a part of it and then 
reviews some of the related works and their applications.

\subsection{Autoencoder (AE)}\label{subsec2}
\begin{figure}[h]
	\centering
	\captionsetup{justification=centering}
	\includegraphics[width=0.4\textwidth]{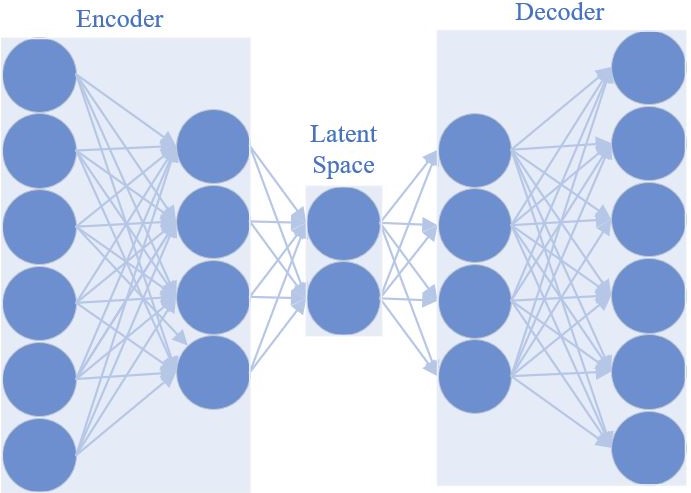}
	\caption{Autoencoder architecture.}
	\label{fig:AE}
\end{figure}

Autoencoders are a type of neural network architecture used for unsupervised learning and dimensionality reduction. The primary goal of an autoencoder is to encode input data into a lower-dimensional representation called the latent space and then decode it back to the original input as accurately as possible. The process of reducing unnecessary correlations or capturing essential features in the latent space can be attributed to the regularization properties of autoencoders [11]. The loss function used during training plays a crucial role. Mean Squared Error (MSE) is a common loss function for autoencoders. It measures the difference between the input and the reconstructed output.
By minimizing the reconstruction error, the autoencoder aims to capture the most significant patterns in the data, discarding noise and irrelevant correlations. Since its input and output are the same, it does not need 
any label and is trained in an unsupervised manner. Its simple 
and powerful properties make it extremely suitable for 
modeling short-term normal behavior. As a result, AEs have 
been used as sub-models for anomaly and change point 
detection models in various works with promising results [12, 
13]. There are algorithms deploying AEs as anomaly detection 
models that operate based on reconstruction error after 
training on normal data [14]-[16]. However, they fail to detect 
long-term anomalies since AE models cannot analyze 
information beyond a short local window. Fig.~\ref{fig:AE} shows the 
simple architecture of AE.

\subsection{Transformer}\label{subsubsec2}
\begin{figure}[ht]
	\centering
	\captionsetup{justification=centering}
	\includegraphics[width=0.4\textwidth]{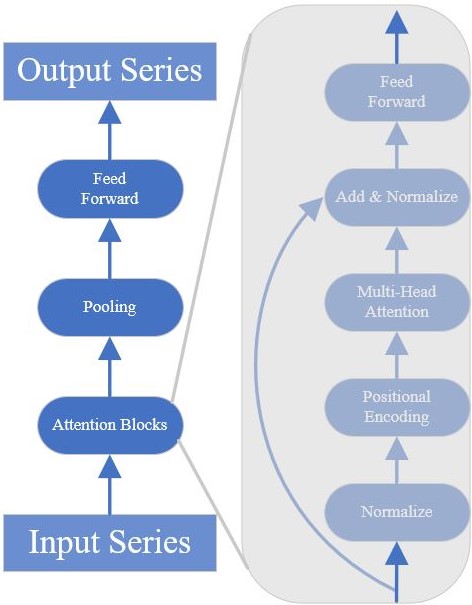}
	\caption{The proposed transformer architecture.}
	\label{fig:Tran.}
\end{figure}

A transformer is a neural network model, which relies on 
attention mechanisms, dispensing with recurrent and 
convolutional networks entirely. This model is highly 
parallelizable, requiring significantly less training and testing 
time. Prior to the introduction of the transformer model, 
recurrent neural networks, such as long short-term memory (LSTM) [17] and gated recurrent unit (GRU) [18] were 
mainly used for sequence modeling. Due to their power in 
sequence modeling and understanding the temporal relations, 
their slow performance was neglected. Their operation is 
based on receiving a data point at each time step as the input 
and generating a sequence of hidden states as a function of 
the previous hidden states and the input. This procedure does 
not easily render itself to parallelization, which is a challenge 
when these models are used for long sequences as time and 
memory become severe constraints. In the transformer model, 
time hierarchy is eliminated. In time series, the order of the 
data points is very important. Hence, transformer uses 
positional embedding to keep track of the order, and at the 
same time, benefits from advantages of parallelization.
The attention-based architecture in our proposed model has 
the same essence as the conventional transformer model, as it 
uses both the positional embedding and the multi-head 
attention. However, it has significant architectural 
differences, such as using a pruned decoder and a pooling 
layer. Fig.~\ref{fig:Tran.} depicts the overall architecture of the 
proposed modified transformer model.
The proposed method in this paper does not consist of any 
recurrent or convolutional neural networks to pave the way 
for parallel computation and benefits from attention 
mechanism to better capture the temporal relations. In the 
following, some of the research works will be discussed.
In Wang et al. [19], an improved LSTM block is proposed 
and its performance in timeseries forecasting is assessed 
compared to LSTM original block and GRU block. Since the 
proposed LSTM block achieves better forecasting 
performance, it is used to calculate the forecasting error and 
detect anomalies with forecasting. The proposed block is 
tested on rail transit operation data.
In Ashraf et al. [20], a neural architecture consisting of 
LSTM blocks in an auto-encoding arrangement is proposed 
which is trained on clean data and detects anomalies based on 
reconstruction error. The network learns allowable patterns in 
the clean data and detects anomalies whenever the 
reconstruction error exceeds a threshold. The proposed model 
is tested on smart transportation data such as in-vehicle 
communication and intrusion are modeled as an anomaly.

\section{The Proposed Method}\label{sec3}
\subsection{Model Operation}\label{subsubsec3}
Let us consider a time series $X = \{x_1, x_2, …, x_n\}$, where $x_i \in \mathbb{R}$ is an $m$-dimensional reading at the $i$th time step that contains information of $m$ channels. The model is trained on widows of the train time series with a size of $W$. The training data should be from normal data with no anomalies. For training, the data is divided into windows of size $W$ starting from the first data point, which gives us $(n-W+1)$ windows of data points. The AE model is trained on all of the windows of data. The weights are adjusted to minimize the mean average error of the output of the decoder. 
The attention model operates in a way to forecast the next step in the latent space of the AE. It is trained on encodings of the AE; in every time step, its input is the encoding of the $t$th window, and its output should be the encoding of the $(t+1)th$ window. The weights are adjusted to minimize the mean average error of the current and the next time-step windows.

In order to achieve better and faster convergence in model training and the ability to use the model for a broader range of datasets, data is normalized. The data is normalized based on the minimum and maximum of the training data. Then, to normalize the rest of the data, we use the same minimum and maximum.
After the training of the model, the function of the model can be summarized as follows:
\begin{itemize}
\item	The window of new data is fed into the encoder of the AE:
\begin{center}
$Encoder(x_1^t, x_2^t, …, x_W^t)$
\end{center}
\item	The encoded window is in the latent space of the AE and is fed to the attention model, and the attention model predicts the encoding of the next step window:
\begin{center}
$Transformer (Encoder(x_1^t, x_2^t, …, x_W^t))$
\end{center}
\item	The prediction is decoded and marked as the model’s prediction of the next time-step window:
\begin{center}
$Prediction=Decoder(Transformer(Encoder(x_1^t, x_2^t, …, x_W^t)))$
\end{center}
\item	Reconstruction error is calculated and appended to the error list:
\begin{center}
$Error = (x_1^{t+1}, x_2^{t+1}, …, x_W^{t+1})$ – $Prediction$
\end{center}
\item	At each time step, the algorithm calculates an average of the error list. If the calculated error is more than average for a few consecutive time steps, it detects the time step as an anomaly.
\end{itemize}

\begin{figure}[ht]
	\centering
	\captionsetup{justification=centering}
	\includegraphics[width=0.6\textwidth]{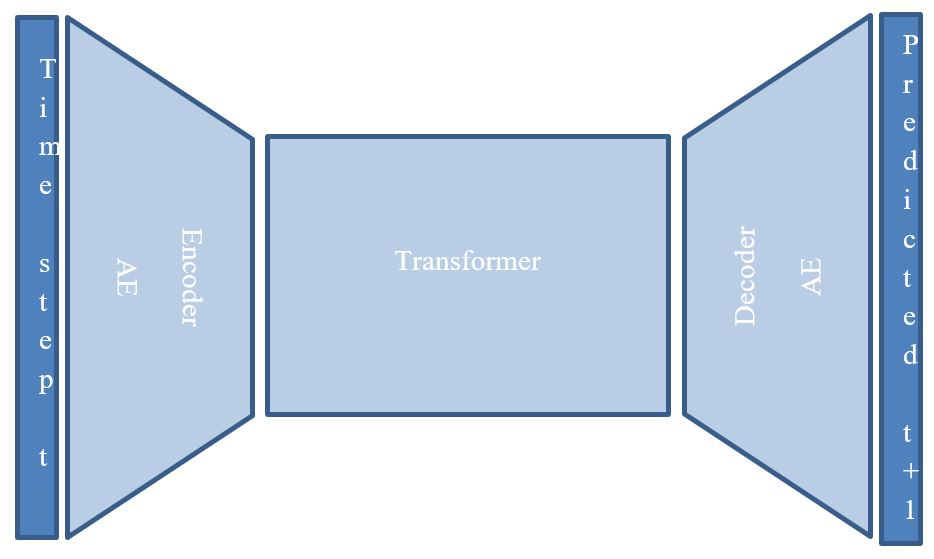}
	\caption{Schematic of the proposed model for anomaly detection.}
	\label{fig:model}
\end{figure}

After training the model, it learns the temporal relationships; therefore, it is expected that if an anomaly from any of the three mentioned types happens in a window, there would be a jump in that window’s reconstruction error for a few consecutive time steps. Fig.~\ref{fig:model} depicts the schematic diagram of the model.

Some anomaly models use a threshold as a criterion for detecting an anomaly [7], but checking if the error is more than average for a few consecutive time steps is more accurate, because this pattern repeats specially for collective and contextual anomalies. Next section covers the implementation details and the obtained results on a number of benchmark time series. 

\subsection{Model Training}\label{subsubsec4}
\subsubsection{Autoencoder Training}\label{subsubsubsec4}
The autoencoder undergoes training on clean data within the primary 
$n$-dimensional space. Given its objective of reconstructing windows, the choice of loss function is pivotal, typically manifesting as either mean squared error (MSE) or mean absolute error (MAE). The autoencoder training process is demonstrated in Algorithm~\ref{autoencoder_training}.
\begin{algorithm}
\caption{Autoencoder Training}\label{autoencoder_training}
\begin{algorithmic}[1]
  \State \textbf{Input:} Time series $X = \{x_1, x_2, …, x_n\}$, window size $W$
  \State \textbf{Preprocess Data:} Normalize $X$ based on training data
  \State \textbf{Initialize:} Autoencoder Model ($\text{AE}$) parameters
  \State \textbf{Training:}
  \For{$t \gets 1$ to $(n-W+1)$}
    \State $window_t \gets (x_t, x_{t+1}, ..., x_{t+W-1})$
    \State $encoded\_window_t \gets \text{AE.Encoder}(window_t)$
    \State $predicted\_window_t \gets \text{AE.Decoder}(encoded\_window_t)$
    \State \textbf{Compute loss and update AE parameters using backpropagation}
  \EndFor
\end{algorithmic}
\end{algorithm}

\subsubsection{Transformer Training}\label{subsubsubsec5}

The improved transformer is subjected to training within the 
$m$-dimensional ($m < n$) latent space of the autoencoder, utilizing clean data. Its primary goal is the prediction of the succeeding window within the latent space of the autoencoder. A comprehensive demonstration of the transformer training process is presented in Algorithm~\ref{transformer_training}.
\begin{algorithm}
\caption{Transformer Training}\label{transformer_training}
\begin{algorithmic}[1]
  \State \textbf{Input:} Time series $X = \{x_1, x_2, …, x_n\}$, window size $W$
  \State \textbf{Preprocess Data:} Normalize $X$ based on training data
  \State \textbf{Initialize:} Autoencoder Model ($\text{AE}$), Transformer Model ($\text{Transformer}$) parameters
  \State \textbf{Training:}
  \For{$t \gets 1$ to $(n-W)$}
    \State $window_t \gets (x_t, x_{t+1}, ..., x_{t+W-1})$
    \State $encoded\_window_t \gets \text{AE.Encoder}(window_t)$
    \State $predicted\_encoded\_window_t \gets \text{Transformer}(encoded\_window_t)$
    \State $target\_encoded\_window_{t+1} \gets \text{AE.Encoder}(x_{t+W})$
    \State \textbf{Compute loss and update Transformer parameters using backpropagation}
  \EndFor
\end{algorithmic}
\end{algorithm}

\section{Results}\label{sec4}

This section presents the experimental results obtained by the proposed model based on six real-world datasets; the internal temperature of an industrial machine, Amazon Web Services (AWS) cloud watch data for Relational Database Service (RDS) and Elastic Compute Cloud (EC2), average CPU usage across a given cluster from AWS CPU usage monitoring, real traffic data (occupancy) from the Twin Cities Metro in Minnesota, and number of Twitter mentions of Google. All of the datasets were retrieved from Numenta Anomaly Benchmark (NAB) [19]. These datasets are carefully chosen to demonstrate the ability of our proposed model in anomaly detection for datasets with various real-world sources as well as all types of anomalies, for instance, contextual and collective anomalies are observed in machine temperature and CPU utilization datasets and point anomaly is observed in AWS EC2 dataset. In the experiments, all of the predicted anomalies, which lie in the dataset's labeled anomaly window as true positives, were counted. 

\begin{table}[h]
	\begin{center}
		\caption{Results achieved by the proposed model for different time series}\label{tab1}%
		\begin{tabular}{@{}lllll@{}}
			\toprule
			Dataset                & Window Size  & Precision & Recall & F1 Score\\
			\midrule
			Machine Temperature    & 90   & 98.0\%  & 100\%  & 99.0\%  \\
			CPU Utilization        & 50   & 100\%  & 100\%  &100\%    \\
			AWS RDS                & 90   & 96.4\%  & 100\% & 98.2\%  \\
			AWS EC2                & 70   & 96.2\%  & 100\% &98.1\%   \\
			Google’s Tweet Volume  & 70   & 93.1\%  & 100\% &96.4\%   \\
			Traffic Occupancy      & 70   & 100\%  & 100\%  & 100\%   \\
			\hline
		\end{tabular}
	\end{center}
\end{table}

\begin{figure}
	\centering
	\begin{subfigure}{0.9\textwidth}
		\centering
		\includegraphics[width=0.85\textwidth]{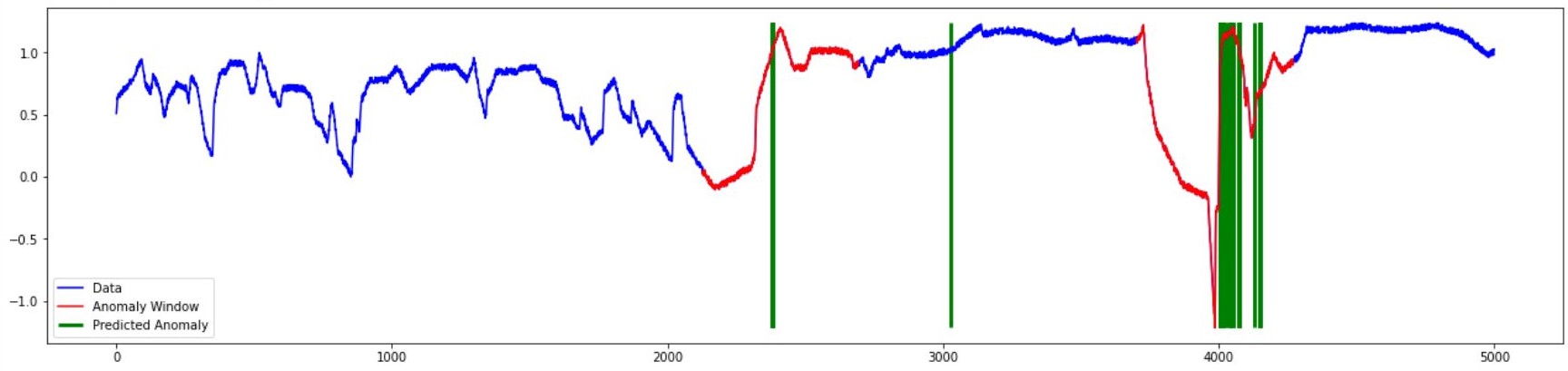}
		\caption{Machine Temperature Dataset}
		\label{fig:y equals x}
	\end{subfigure}
	\begin{subfigure}{0.9\textwidth}
		\centering
		\includegraphics[width=0.85\textwidth]{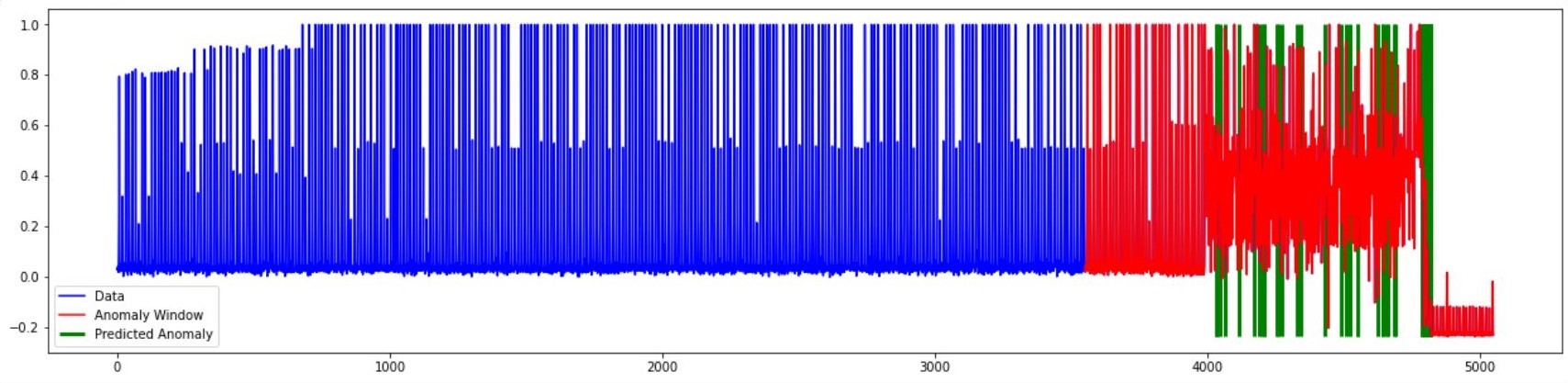}
		\caption{CPU Utilization Dataset}
		\label{fig:five over x}
	\end{subfigure}
	\begin{subfigure}{0.9\textwidth}
		\centering
		\includegraphics[width=0.85\textwidth]{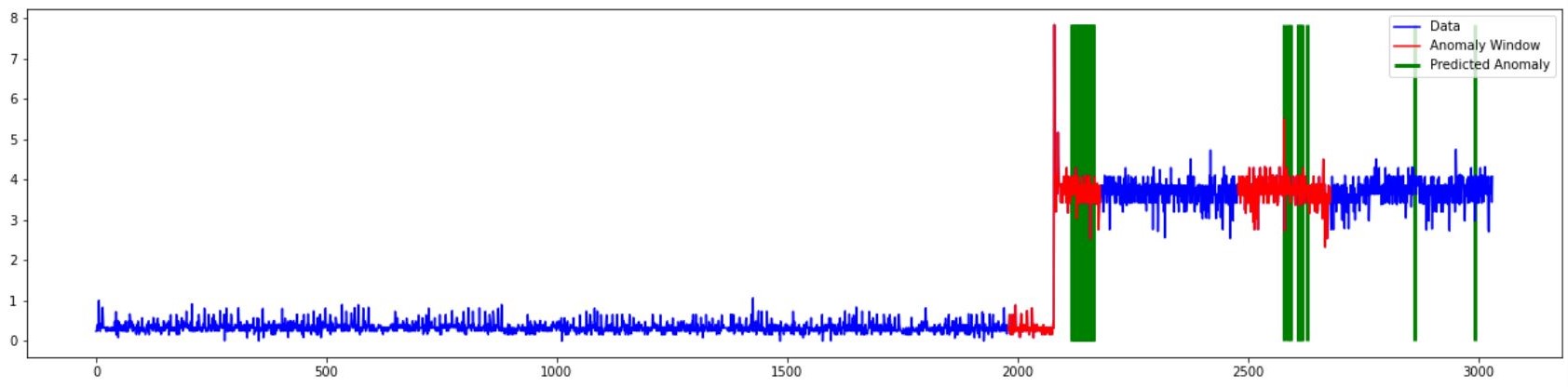}
		\caption{AWS RDS Dataset}
		\label{fig:three sin x}
	\end{subfigure}
	\begin{subfigure}{0.9\textwidth}
		\centering
		\includegraphics[width=0.85\textwidth]{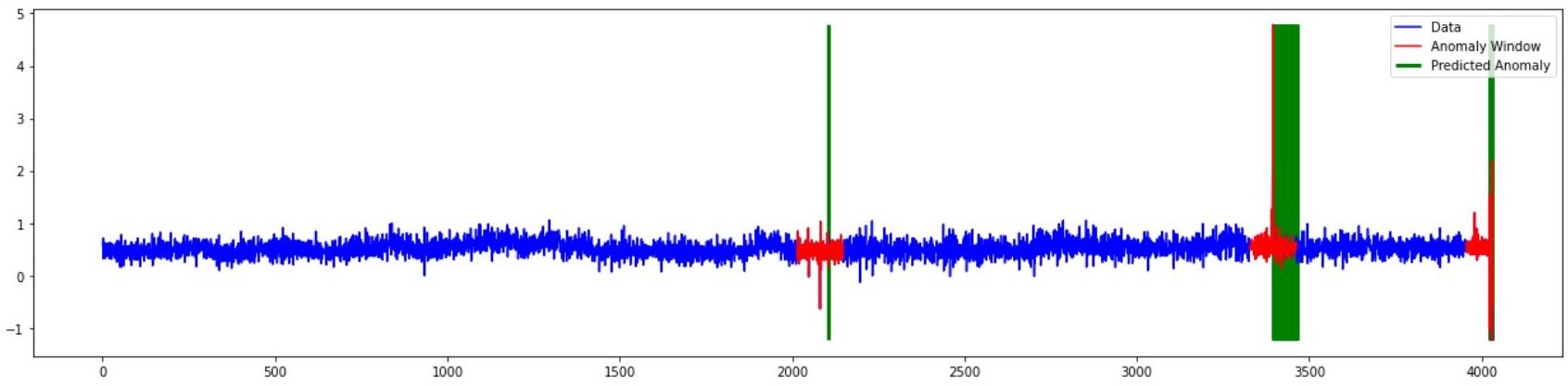}
		\caption{AWS EC2 Dataset}
		\label{fig:five over x}
	\end{subfigure}
	\begin{subfigure}{0.9\textwidth}
		\centering
		\includegraphics[width=0.85\textwidth]{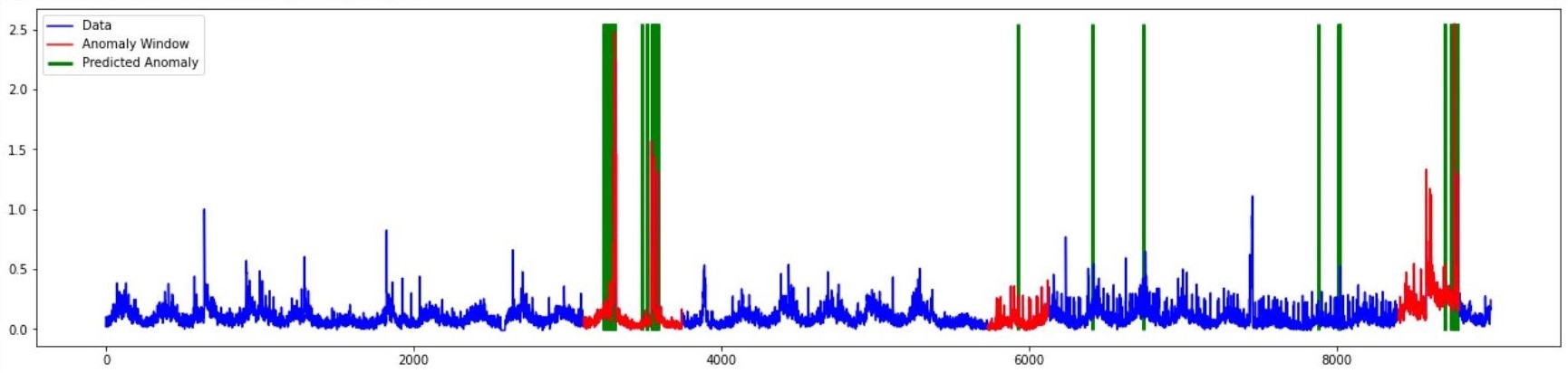}
		\caption{Google's Tweet Volume Dataset}
		\label{fig:five over x}
	\end{subfigure}
	\begin{subfigure}{0.9\columnwidth}
		\centering
		\includegraphics[width=0.85\textwidth]{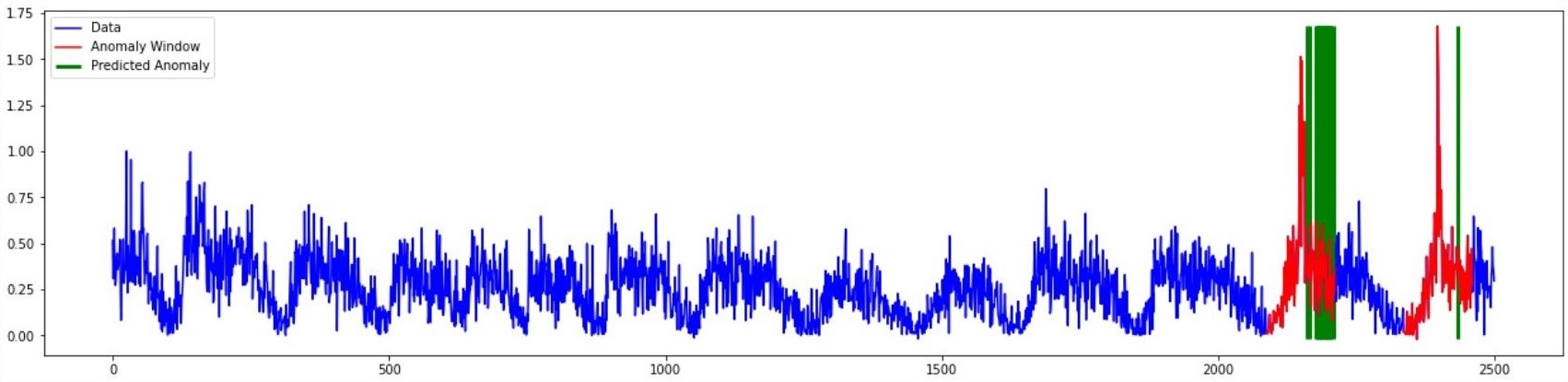}
		\caption{Traffic Occupancy Dataset}
		\label{fig:five over x}
	\end{subfigure}
	\caption{Experimental results for different time series.}
	\label{fig:results}
\end{figure}

In the experiments, two anomaly detection algorithms were used and evaluated based on their reconstruction errors. The first algorithm is based on the pattern, which is repeated at every anomaly. It is more reliable for contextual and collective anomalies and is able to operate even without validation data. It checks if the difference between the calculated errors in the last two steps is more than the average reconstruction error in the previous steps. The preceision of this algorithm is increased as it is used for more consecutive error pairs. This algorithm is used for anomaly detection in machine temperature, CPU utilization, traffic occupancy, and AWS RDS datasets. Since anomalies in these datasets are mostly contextual or collective.
The second algorithm uses a specific number determined by validation data as a threshold to announce a point as an anomaly, if its reconstruction error is more than the specified threshold. This method needs validation data to determine the threshold and works best for point anomalies, and has a weaker performance in detecting contextual and collective anomaly types.
Table~\ref{tab1} demonstrates the accuracy achieved by the proposed algorithm for different time series using the mentioned prediction window size. Fig.~\ref{fig:results} demonstrates the proposed model's predictions, anomaly points, and anomaly windows labeled in the datasets.

According to Table~\ref{tab1} and Fig.~\ref{fig:results}, the proposed model shows promising results in detecting all three types of anomalies. Moreover, most of the false positives announced by our model have considerable differences from other normal points in the time series. However, they are not labeled as anomalies in the datasets. Our model detects all of the anomalies in all six datasets and performs with the highest precision in two of them. Its lowest precision occurs in Google's Tweet Volume Dataset, which is 93.1\%. The reason might be the anomalous patterns that are repeated throughout this dataset. Table~\ref{tab2} compares the proposed model with four other well-established deep models. The proposed model outperforms all of them for two of the datasets with a wide margin, while losing to only one of them for the third dataset with a narrow margin of 1.5\% in F1 Score.

\begin{table}[h]
	\begin{center}
		\caption{Comparison between different anomaly detection algorithms applied to different time series}\label{tab2}%
		\begin{tabular}{@{}llllll@{}}
			\toprule
			Dataset                & Model & Window Size  & Precision & Recall & F1 Score\\
			\midrule
			
			& Proposed method         & 90   & 98.0\%      & \bf{100\%}  & \bf{99.0\%}  \\
			& VAE-LSTM [7] & 288   & 55.9\%     & \bf{100\%}  &71.7\%    \\
			Machine Temperature & VAE [20]     & 48   & 21.1\%      & \bf{100\%}  &20.7\%    \\
			& LSTM-AD [21] & 48   & \bf{100\%}  & 50.0\%       &66.7\%    \\
			& ARMA [22]    & 48   & 14.2\%       & \bf{100\%}  &24.8\%    \\
			
			\hline
			
			& Proposed method         & 50   & \bf{100\%}  & \bf{100\%}  & \bf{100\%}  \\
			& VAE-LSTM [7] & 144  & 69.4\%     & \bf{100\%}  &81.9\%    \\
			CPU Utilization		 & VAE [20]     & 24   & 34.8\%     & 50.0\%       &41.0\%    \\
			& LSTM-AD [21] & 24   & 27.4\%     & \bf{100\%}  &43.0\%    \\
			& ARMA [22]    & 24   & 23.4\%     & \bf{100\%}  &38.0\%    \\
			
			\hline
			
			& Proposed method         & 70   & 96.2\%      & \bf{100\%}  & 98.1\%  \\
			& VAE-LSTM [7] & 192   & 99.3\%     & \bf{100\%}  &\bf{99.6\%}    \\
			AWS EC2				 & VAE [20]     & 24   & 94.9\%      & \bf{100\%}  &97.4\%    \\
			& LSTM-AD [21] & 24   & \bf{100\%}  & 43.6\%      &60.8\%    \\
			& ARMA [22]    & 24   & 93.8\%      & \bf{100\%}  &96.8\%    \\
			
			\hline
		\end{tabular}
	\end{center}
\end{table}
Because of its unique architecture and absence of recurrent
and convolutional layers in our proposed model, it is highly 
computationally parallelizable. Therefore, deploying a GPU in 
training or employing it would boost its performance and make it a suitable choice for real-time applications. As a testify, in training the model on EC2 dataset with 1700 time 
steps of normal training data using Google Colab’s Tesla T4 
GPU, each epoch of autoencoder training roughly takes 0.15 
seconds on GPU while taking 0.18 seconds without GPU. The 
gap is way more significant for training the attention-based 
model, as it takes about 20 seconds for each epoch on GPU 
and takes about 860 seconds in training without GPU. The 
autoencoder converges after about 500 epochs, while the 
attention-based model needs only five epochs to converge.

\section{Conclusion}\label{sec5}
In this work, we proposed an attention-based and 
autoencoder hybrid model as a solution for unsupervised 
anomaly detection in time series. Our proposed method 
benefits from an autoencoder to learn the local features and to 
summarize the windows into short embeddings used by the 
attention-based model as input. The attention-based model is 
built on the transformer model [8] with major architectural 
differences, such as deploying a pooling layer and a pruned 
decoder. The attention-based architecture is used to model 
long-term temporal relations. The proposed model does not 
use recurrent or convolutional layers. Hence, it renders itself 
to parallel computation, which in turn expedites the training 
process and reduces the computation time in using the trained 
model. This paves the way for using the proposed algorithm in 
real-time tasks. The proposed algorithm was applied to six 
different real-world datasets from the NAB anomaly detection 
benchmark, and its performance was compared with five well-established models. Results confirmed the effectiveness of the 
proposed architecture for unsupervised online anomaly 
detection.

\bibliographystyle{plain}

\end{document}